\documentclass{article}
\usepackage{ijcai24}

\usepackage{times}
\usepackage{soul}
\usepackage{url}
\usepackage[hidelinks]{hyperref}
\usepackage[utf8]{inputenc}
\usepackage[small]{caption}
\usepackage{graphicx}
\usepackage{amsmath}
\usepackage{amsfonts}
\usepackage{amsthm}
\usepackage{booktabs}
\usepackage{algorithm}
\usepackage{algorithmic}
\usepackage[switch]{lineno}
\usepackage{xcolor}
\usepackage{multirow}
\usepackage{bm}

\title{CPT: Competence-progressive Training Strategy for Few-shot Node Classification}

\author{
    Qilong Yan$^1$, 
    Yufeng Zhang$^3$, 
    Jinghao Zhang$^3$, 
    Jingpu Duan$^2$, 
    Jian Yin$^1$
    \affiliations
    $^1$Sun Yat-sen University\\
    $^2$Peng Cheng Laboratory\\
    $^3$Institute of Automation, Chinese Academy of Sciences
    \emails
     yanqlong@mail2.sysu.edu.cn, 
     yufeng.zhang@ia.ac.cn, 
     jinghao.zhang@cripac.ia.ac.cn, \\
     duanjp@pcl.ac.cn, 
     issjyin@mail.sysu.edu.cn
}

\begin{document}

\maketitle

\begin{abstract}
Traditional episodic meta-learning methods have made significant progress in solving the problem of few-shot node classification on graph data. However, they suffer from an inherent limitation: the high randomness in task assignment often traps the model in suboptimal solutions. This randomness leads the model to encounter complex tasks too early during training, when it still lacks the necessary capability to handle these tasks effectively and acquire useful information. This issue hampers the model’s learning efficiency and significantly impacts the overall classification performance, leading to results that fall well below expectations in data-scarce scenarios. Inspired by human learning, we propose a model that starts with simple concepts and gradually advances to more complex tasks, which facilitates more effective learning. We introduce CPT, a novel curriculum learning strategy designed for few-shot node classification. This approach progressively aligns the meta-learner’s capabilities with both node and task difficulty, enhancing the learning efficiency for challenging tasks and further improving overall performance. Specifically, CPT consists of two stages: the first stage begins with basic tasks, aligning the difficulty of nodes with the model's capability to ensure that the model has a solid foundational ability when facing more complex tasks. The second stage focuses on aligning the difficulty of meta-tasks with the model's capability and dynamically adjusting task difficulty based on the model's growing capability to achieve optimal knowledge acquisition. Extensive experiments on widely used node classification datasets demonstrate that our method significantly outperforms existing approaches. 
\end{abstract}

\section{Introduction}

In recent years, there has been a lot of research focused on node classification, which is about predicting labels for unlabeled nodes in a graph. This task has many practical applications in real-world situations. For example, it can help us identify molecules that play a crucial role in protein structures \cite{szklarczyk2019string}, recognize important influencer in social networks \cite{yuan2013latent,zhang2020every}, or infer users' demographic information based on behavior patterns in recommendation systems \cite{yan2021relation}.
Although Graph Neural Networks (GNNs) have been successful in node classification tasks \cite{Wu_Pan_Chen_Long_Zhang_Yu_2021,xu2018powerful}, their performance heavily relies on the number of labeled nodes in each class \cite{Fan_Chengtai_Kunpeng_Goce_Ting_Ji_2019}.
 In real-world graphs, we often have a large number of labeled nodes in some classes, while other classes have only a limited number of labeled nodes. We refer to classes with a large number of labeled samples as  \textit{base classes}, and those with a limited number of labels as \textit{novel classes} \cite{wang2022task}. For example, in disease diagnosis, common diseases with abundant samples can be considered base classes, while newly discovered diseases with limited samples are regarded as novel classes. Due to the widespread existence of novel classes in real-world graphs, it is important for GNNs to be able to perform node classification with a limited amount of labeled nodes, a task known as few-shot node classification.

\begin{figure}[t]
  \includegraphics[width=0.47\textwidth]{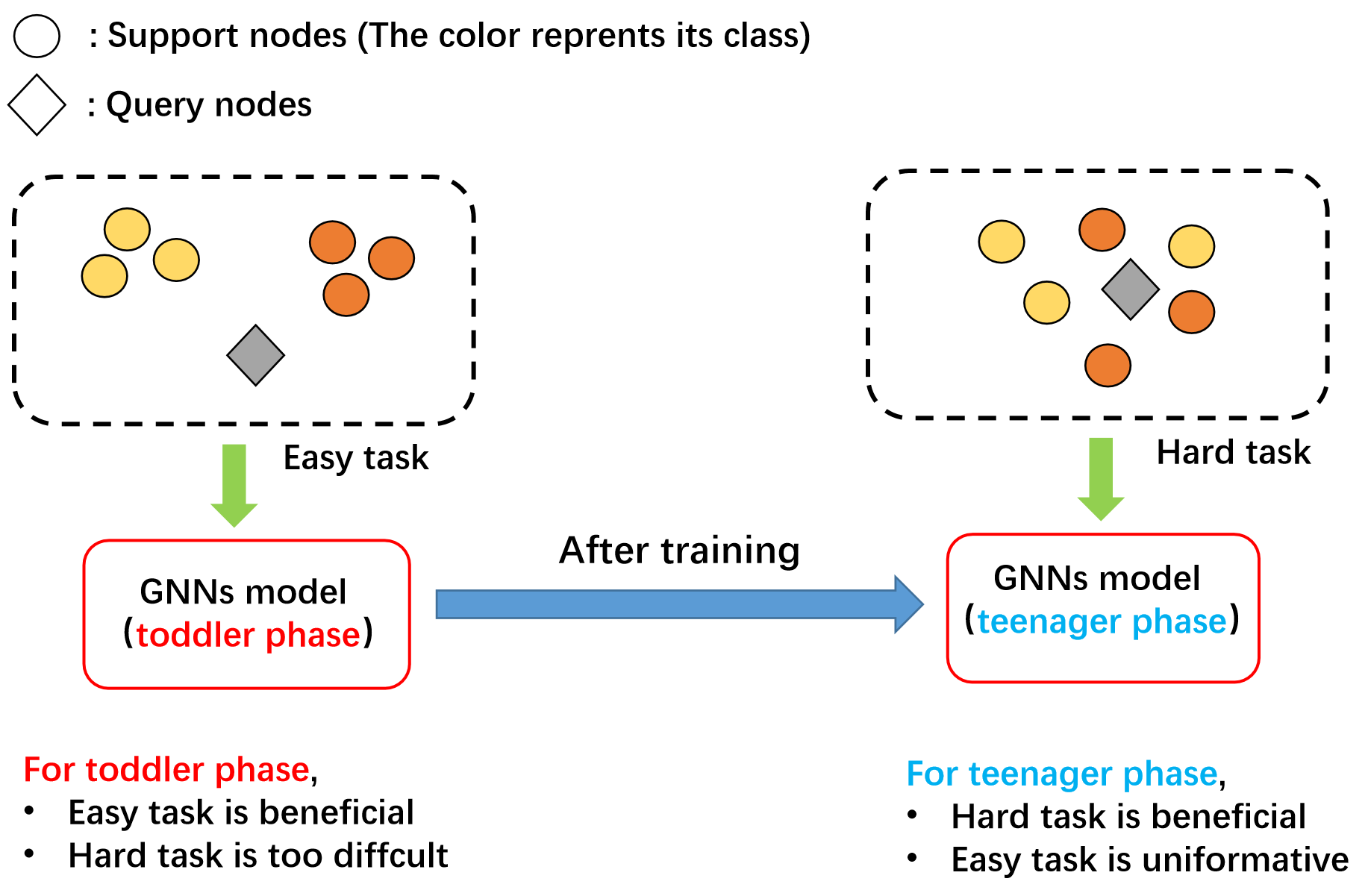} 
  \caption{ \textbf{ Tasks influence a model's intelligence differently depending on its growth stage, mirroring human learning patterns.}}
  \label{fig:intro}
  \vspace{-5mm}
\end{figure}

\begin{figure}[t]
  \includegraphics[width=0.45\textwidth,keepaspectratio]{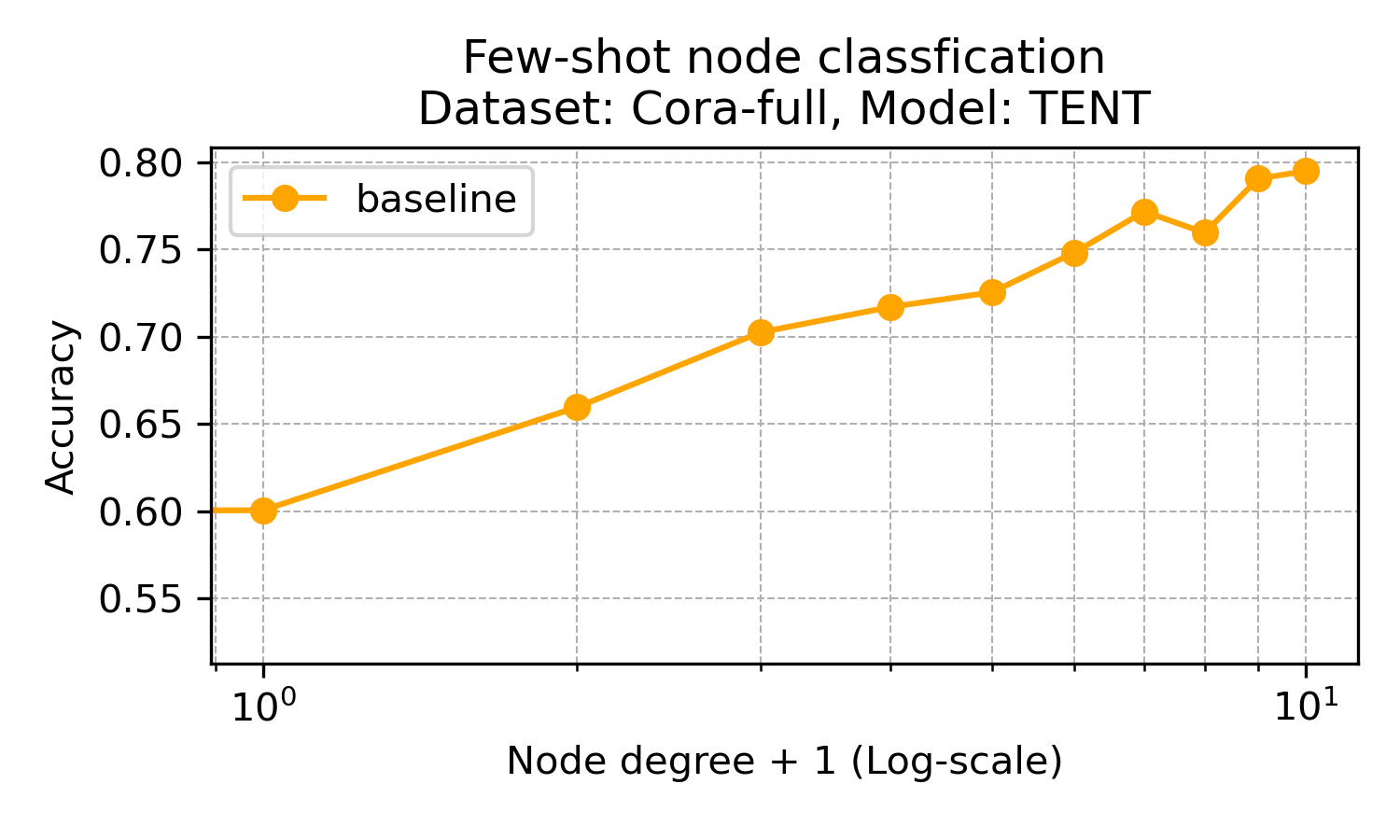} 
    \vspace{-5mm}
  \caption{ \textbf{ Degree-specific generalization performance of
the few-shot node classification task. The x-axis represents the node degrees in the training graph and the y-axis is the generalization performance averaged over nodes with the specific degrees.}}
  \label{fig:degree}
    \vspace{-2mm}
\end{figure}

Unlike the traditional node classification problem, few-shot node classification has training data with only a small number of samples, which leads to the challenge of insufficient knowledge. The current mainstream learning paradigm addresses this issue by constructing meta-tasks through meta-learning to transfer knowledge \cite{wang2022task,Wang_Luo_Ding_Zhang_Li_Zheng_2020,Liu_Fang_Liu_Hoi_2021}.
 In this paradigm, the meta-learner learns from a series of meta-training tasks derived from the base classes and evaluates the model using meta-testing tasks derived from the novel classes. The goal of this approach is to enable the meta-learner to extract meta-knowledge from the base classes, allowing it to classify novel classes with only a few labeled nodes. Although this meta-learning paradigm has achieved success in the few-shot node classification task, it has inherent flaws in its context-based training strategy. This strategy randomly selects meta-tasks for learning, which may lead to certain issues. In node classification tasks, one of the primary factors affecting overall performance is learning from difficult nodes, and a similar situation exists in few-shot graph learning, as shown in Figure \ref{fig:degree}. Since meta-tasks are composed of nodes in meta-learning, there are also differences in task difficulty. The context-based training strategy randomly selects tasks from the base classes for learning, ignoring the differences in task difficulty. This may cause a mismatch between the meta-learner’s capability and the task’s difficulty, leading to suboptimal performance when learning from difficult tasks and thereby affecting overall results.

 This idea is inspired by the human learning process, where humans typically start learning from basic concepts and gradually progress to more complex content, as harder tasks require knowledge from simpler ones as a foundation. This strategy of progressively learning from easy to hard tasks has been effectively applied in curriculum learning \cite{Hacohen_Weinshall_2019} and adaptive learning \cite{Jiang_Meng_Yu_Lan_Shan_Hauptmann_2014}. In these frameworks, individual samples are assigned different levels of importance (i.e., weights) to enhance the model’s generalization ability. In other words, randomly selecting tasks may expose the meta-learner to high-difficulty tasks too early, resulting in ineffective acquisition of useful information from those challenging tasks. As shown in the Figure \ref{fig:intro}, at the initial stage (analogous to an infant stage), the model lacks the capability to handle difficult tasks but can acquire knowledge from simpler tasks. At this stage, simple tasks are rich in information and beneficial to the model. After sufficient training, the model’s intelligence level improves, resembling an adolescent stage, where learning from difficult tasks becomes more effective due to its foundational capabilities. While previous studies have explored the use of curriculum learning in node classification tasks \cite{wei2023clnode}, their approaches are primarily focused on nodes and cannot be applied to meta-tasks, such as node-level oversampling or reparameterization. Although there are curriculum learning approaches designed for meta-tasks \cite{Sun_Liu_Chua_Schiele_2019,Liu_Wang_Sahoo_Fang_Zhang_Hoi_2020}, they focus on image and text data, without considering the complex relational dependencies in graph data. Therefore, new solutions are needed to address the challenges of few-shot classification on graph data.  

  In this paper, we propose \textbf{C}ompetence-\textbf{P}rogressive \textbf{T}raining strategy for few-shot node classification, named as CPT. CPT aims to provide tasks of different difficulty levels for meta-learners at different stages to adapt to their current abilities and alleviate the issue of suboptimal points. Specifically, CPT uses the two-stage strategy \cite{Bengio_Louradour_Collobert_Weston_2009,hu2022tuneup} to train a meta-learner. In the first stage, CPT follows previous research by randomly sampling nodes to construct meta tasks and then employs the default training strategy(meta training), i.e., simply minimizing the given supervised loss, to produce a strong base meta-learner to start with. Although the base meta-learner may perform well on easy tasks and has gained some knowledge, it no longer benefits or improves its abilities from these tasks, and it might converge to suboptimal points. So the second stage focuses on generating more difficult tasks and fine-tuning it. Specifically, CPT increases the difficulty of each task by dropping the edges of its graph. The underlying motivation for this approach is to create a larger number of 'tail nodes' within the meta-training tasks. This alteration is intended to enrich the meta-learning model with additional 'tail node' meta-knowledge, as these nodes, with their fewer connections, typically present unique challenges in learning processes. Finally, CPT finetunes the base meta-learner by minimizing the loss over the increased hard task data. CPT can be easily implemented in the existing meta-learner training pipeline on graphs, as demonstrated in Algorithm  \ref{alg:1}. Additionally, CPT is compatible with any GNN-based meta-learner and can be employed with any desired supervised loss function.

Our main contributions can be summarized as follows:

\begin{itemize}
    \item \textbf{Problem.} We explore the limitations of context-based meta-learning approaches and discuss their relevant issues in few-shot node classification. Specifically, the limitation of these approaches lies in their random and uniform sampling of tasks, which may result in insufficient training of the meta-learner on difficult tasks, leading to suboptimal model performance.
    \item \textbf{Method.} We present CPT, a novel two-stage curriculum learning strategy for few-shot node classification. CPT is the first method to solve such problems in few-shot tasks on graphs. Notably, this is a general strategy that can be applied to any GNN architecture and loss function.
    \item \textbf{Experiments.} We conducted experiments on four benchmark node classification datasets under few-shot settings, and the results demonstrate that our strategy significantly improves performance over existing methods.
\end{itemize}

\section{PRELIMINARIES}

\subsection{Problem Definition}
In this work, we focus on few-shot node classification (FSNC) tasks on a single graph. Formally, let $G=(\mathcal{V},\mathcal{E}, X)=(A,X)$ denote an attributed graph, where $A=\mathop{\{0, 1\}}^{V \times V}$ is the adjacency matrix representing the network structure, $X = [x_1; x_2; ...; x_V]$ represents all the node features, $\mathcal{V}$ denotes the set of vertices $\{v_1, v_2, ..., v_V\}$, and $\mathcal{E}$ denotes the set of edges $\{e_1, e_2, ..., e_E\}$. Specifically, $A_{j,k} = 1$ indicates that there is an edge between node $v_j$ and node $v_k$; otherwise, $A_{j,k} = 0$. The few-shot node classification problem assumes the existence of a series of node classification tasks, $\mathcal{T} = \{T_i\}_{i=1}^{I}$, where $T_i$ denotes the given dataset of a task, and $I$ denotes the number of such tasks. FSNC tasks assume that these tasks are formed from target novel classes (i.e., $\mathcal{C}_{\text{novel}}$), where only a few labeled nodes are available per class. Additionally, there exists a disjoint set of base classes (i.e., $\mathcal{C}_{\text{base}}$, $\mathcal{C}_{\text{base}} \cap \mathcal{C}_{\text{novel}} = \emptyset$) on the graph, where the  number of labeled nodes is sufficient. For better understanding, we first present the definition of an N-way K-shot node classification task as follows:

\textsc{Definition 1.} \textit{\textbf{N-way K-shot Node Classification:}  
Given an attributed graph $ G = (A, X)$ with a specified node label space $\mathcal{C}$, $|\mathcal{C}| = N$. If for each class $c \in \mathcal{C}$, there are K labeled nodes (i.e., support set $\mathcal{S}$) as references and another R nodes (i.e., query set $\mathcal{Q}$) for prediction, then we term this task as an N-way K-shot Node Classification task.}

\textsc{Definition 2.} \textit{\textbf{Few-shot Node Classification:} 
Given an attributed graph $G = (A, X)$ with a disjoint node label space $\mathcal{C} = \{\mathcal{C}_{\text{base}}, \mathcal{C}_{\text{novel}}\}$. Substantial labeled nodes from $\mathcal{C}_{\text{base}}$ are available for sampling an arbitrary number of $N$-way $K$-shot Node Classification tasks for training. The goal is to perform $N$-way $K$-shot Node Classification for tasks sampled from $\mathcal{C}_{\text{novel}}$.}

Few-shot Node Classification extends N-way K-shot Node Classification by utilizing labels from base classes for task training, while evaluation is performed using labels from novel classes. This approach assesses the model's capacity to generalize and predict previously unseen categories.

\subsection{Meta-Training on Graph}
The meta-training and meta-test processes are conducted on a certain number of meta-training tasks and meta-test tasks, respectively. These meta-tasks share a similar structure, except that meta-training tasks are sampled from $C_b$, while meta-test tasks are sampled from $C_n$. The main idea of few-shot node classification is to keep the consistency between meta-training and meta-test to improve the generalization performance.

To construct a meta-training (or meta-test) task $ \mathcal{T}_t$, we first randomly sample $N$ classes from $C_b$ (or $C_n$). Then we randomly sample $K$ nodes from each of the $N$ classes (i.e., $N$-way $K$-shot) to establish the support set $\mathcal{S}_t$. Similarly, the query set $Q_t$ consists of $Q$ different nodes (distinct from $\mathcal{S}_t$) from the same $N$ classes. The components of the sampled meta-task $\mathcal{T}_t$ can be denoted as follows:

\begin{align*}\label{2}
    & \mathcal{S}_t= \{ (v_1,y_1), (v_2,y_2),..., (v_{N \times K},y_{N \times K}) \}, \\
    \vspace{10mm}
    & Q_t= \{ (q_1,y^{'}_1), (q_2,y_2^{'}),..., (q_{Q},y_{Q}^{'}) \}, \\
    \vspace{10mm}
    & \mathcal{T}_t= \{ \mathcal{S}_t, Q_t \},
    \vspace{10mm}
\end{align*}

where $v_i$ (or $q_i$) is a node in $\mathcal{V}$, and $y_i$ (or $y_i^{'}$) is the corresponding label. In this way, the whole training process is conducted on a set of $T$ meta-training tasks $\mathcal{T}_{train} = \{\mathcal{T}_t\}^T_{t=1}$. After training, the model has learned the transferable knowledge from $\mathcal{T}_{train}$ and will generalize it to meta-test tasks $\mathcal{T}_{test} = \{\mathcal{T}_t^{'}\}^{T_{test}}_{t=1}$ sampled from $C_n$.
\vspace{2mm}


\section{METHODOLOGY}
In this section, we introduce CPT which uses a curriculum learning strategy to better train a Graph Neural Network meta-learner (GNN meta-learner) in a few-shot node classification problem. 
In general, CPT applies the two-stage strategy to train a GNN meta-learner. The first stage aims to produce a strong base GNN meta-learner that performs well on easy tasks. In the first stage, the meta-learner has acquired knowledge of easy tasks and gained some preliminary abilities, enabling it to tackle more challenging tasks. Then, in the second stage, the meta-learner focuses on increasing the difficulty of tasks, progressively learning more complex tasks based on its growing competencies. This strategy allows the meta-learner to gradually adapt and master increasingly challenging tasks.

\subsection{First stage: Default meta-training.}

In the first stage, we aim to obtain a base meta-learner $f_\theta$ that has some preliminary abilities, enabling it to tackle more challenging tasks in the second stage. Specifically, we use the episodic meta-learning scheme to converge the GNN meta-learner on foundational tasks. These foundational tasks are considered easy because the tasks in the second stage are designed to increase in difficulty based on them. Next, we calculate the loss based on the model's loss function. For the sake of brevity, we employ the cross-entropy loss as the loss function for the model when considering the base classes $C_b$. Formally, this can be expressed as follows:
\vspace{-1mm}
\begin{equation}
    Z_i = {\rm softmax}\left(f_\theta(h_i)\right)
\end{equation}
\begin{equation}
    \mathcal{L} = -\sum^{Q}_{i=1}\limits\sum\limits^{|C_b|}_{j=1} Y_{i,j}{\rm log}(Z_{i,j})
\end{equation}
where $Z_i$ is the probability that the $i$-th query node in $Q$ belongs to each class in $C_b$. $Y_{i,j}=1$ if the $i$-th node belongs to the $j$-th class, and $Y_{i,j}=0$, otherwise. $Z_{i,j}$ is the $j$-th element in $Z_i$. Then we update meta-learner parameters $\theta$ via one gradient descent step in task $\mathcal{T}_i$:
\begin{equation}
    \theta^{'} = \theta - \alpha_1 \frac{\partial \mathcal{L}_{\mathcal{T}_i}(f_\theta)}{\partial \theta}       
\end{equation}
where $\alpha_1$ is the task-learning rate and the model parameters are trained to optimize the performance of $f_{\theta^{'}}$ across meta-training tasks. The model parameters $\theta$ are updated as follows:
\begin{equation}
    \theta = \theta - \alpha_2 \frac{\partial \sum_{\mathcal{T}_i \sim p(\mathcal{T}_{train})} \mathcal{L}_{\mathcal{T}_i}(f_{\theta^{'}_i})}{\partial \theta}       
\end{equation}
where $\alpha_2$ s the meta-learning rate, $p(\mathcal{T}_{train})$ is the distribution of meta-training tasks.
The detailed training procedure is described in $L2-8 $ of Algorithm \ref{alg:1}.

\subsection{Second stage: Competence-progressive training.}

After the first stage, although the meta-learner performs well on foundational tasks, it is prone to converging to suboptimal points.
To solve this issue, we generate increasingly hard tasks to help the meta-learner break through suboptimal points. 
There are many possible ways of controlling the difficulty of meta tasks in graph data. In our approach, we utilize the degree of nodes as the key factor to modulate the complexity of these tasks.
Node degree is the number of immediate neighbors of a node in a graph. Our investigations have uncovered an implicit correlation between the degree of nodes and their performance. specifically, nodes with fewer connections—termed 'tail nodes'— tend to present greater challenges in learning processes\cite{liu2021tail}. We have observed a similar phenomenon in few-shot node classification tasks, as depicted in Figure \ref{fig:degree}. We can see that the few-shot node classification also faces performance issues on tail nodes. So, we use the DropEdge methods \cite{Rong_Huang_Xu_Huang_2019} to randomly delete edges from the entire graph. 
The underlying motivation for this approach is to create a larger number of 'tail nodes' within the meta-training tasks to control task difficulty.
The DropEdge ratio $\beta$ is determined by the model's competence $c$, where $\beta$ = $c$. We employ the following function \cite{Platanios_Stretcu_Neubig_Póczos_Mitchell_2019,vakil2023curriculum} to quantify model's competence:

\begin{equation}
    c(t) = min \left(  1,\sqrt[p]{ t\left( \frac{1-c_0^p}{T} \right) + c_0^p}  \right)
\end{equation}

where $t$ is the current training iteration, $p$ controls the sharpness of the curriculum so that more time is spent on the examples added later in the training, $T$ is the maximum number of iterations, and $c_0$ is the initial value of the competence. $p$ and $c_0$ are manually controlled hyper-parameters.
We can find that the model's competence $c$ increases over time, and the DropEdge ratio $\beta$ also correspondingly rises, leading the model to face harder tasks. The detailed training procedure is described in $L9-17 $ of Algorithm \ref{alg:1}.

Notably, CPT is a curriculum-based learning strategy designed to train any model architecture using any supervised loss to acquire a better meta-learner over graphs, which only adds a few simple components.

\begin{algorithm}[t!]

  \caption{Detailed learning process of CPT. }

    \label{alg:1}
       \textbf{Input:} graph $G = (\mathcal{V},\mathcal{E},\mathcal{X})$, a meta-test task $\mathcal{T}_{test}=\{\mathcal{S},Q\}$,  meta-learner $f_\theta$ , base classess $C_b$, meta-training epochs $T$, The number of classes $N$ , and the number of labeled nodes for each class $K$, DropEdge ratio $\beta$, task-learning rate $\alpha_1$, meta-learning rate $\alpha_2$.       \newline
       \textbf{Output:} Predicted labels of the query nodes $Q$ 

  \begin{algorithmic}[1]
    \STATE \texttt{// Meta-training process}
    \STATE{\textbf{\# First stage: Default meta-training to obtain a base meta-learner.}}

    \FOR{$t=1,2,...,T$ }

     \STATE Sample a meta-training task  $\mathcal{T}_i=\{\mathcal{S}_i,Q_i \}$ from $C_b$;
     \STATE Use $f_\theta$ to Compute node representations for the nodes in set $\mathcal{S}_i$ and  $Q_i$  ;
 
     \STATE Compute the meta-training loss according to the loss function defined by the model;
     \STATE Update the model parameters using the meta-training loss via one gradient descent step, according to Eq. (4) and (5);
     \ENDFOR

    \STATE   \textcolor{red}{\textbf{ \# Second stage: Competence-progressive training for the base meta-learner using increasingly hard tasks.}}

    \FOR{$t=1,2,...,T$ }

     \STATE  $c(t) \gets $ competence from Eq. (6); 
     \STATE DropEdge ratio $\beta = $ $c(t)$;  
     \STATE \textbf{Generate hard task,\textit{i.e.}, randomly drop $\beta$ of edges : $G \stackrel{DropEdge}{\longrightarrow}\widetilde{G}$};  
     \STATE Sample a meta-training task  $\mathcal{T}_i=\{\mathcal{S}_i,Q_i \}$ from $C_b$;
     \STATE Use $f_\theta$ to Compute node representations for the nodes in set $\mathcal{S}_i$ and  $Q_i$
     \STATE Compute the meta-training loss according to the loss function defined by the model;
     \STATE Update the model parameters using the meta-training loss via one gradient descent step, according to Eq. (4) and (5);
     
    \ENDFOR
    \STATE \texttt{// Meta-test process}
     \STATE Use $f_\theta$ to Compute node representations for the nodes in set $\mathcal{S}$ and  $Q$
    \STATE Predict labels for the nodes in the query set $Q$ ;
    
  \end{algorithmic}

\end{algorithm}

\section{EXPERIMENTS}

In this section, we conduct a comprehensive performance evaluation of CPT. Additionally, through experimental observations, we find that CPT can effectively mitigate the suboptimal solution problem. Through ablation experiments, we verify the effectiveness of both the motivation behind CPT and its various components.

\subsection{Datasets}

To evaluate our framework on few-shot node classification tasks, we conduct experiments on four prevalent real-world graph datasets:Amazon-E \cite{McAuley_Pandey_Leskovec_2015}, DBLP \cite{Tang_Zhang_Yao_Li_Zhang_Su_2008}, Cora-full \cite{Bojchevski_Günnemann_2017}, and OGBN-arxiv \cite{Hu_Fey_Zitnik_Dong_Ren_Liu_Catasta_Leskovec_2020}. We summarize the detailed statistics of these datasets in Table \ref{tab:1}. Specifically,  Nodes and  Edges denote the number of nodes and edges in the graph, respectively. Features denotes the dimension of node features. Class Split denotes the number of classes used for meta-training/validation/meta-test. This selection aligns with the default settings established for the respective dataset.

\begin{table}[ht]
\centering

\renewcommand{\arraystretch}{1.5}
\caption{\bf Statistics of four node classification datasets.}
\vspace{-1mm}
\resizebox{0.47\textwidth}{!}{
\begin{tabular}{c|c|c|c|c}
\hline

\textbf{Dataset} & Nodes   & Edges     & Features & Class Split \\ \hline
Amazon-E         & 42,318  & 43,556    & 8,669    & 90/37/40    \\
DBLP             & 40,672  & 288,270   & 7,202    & 80/27/30    \\
Cora-full        & 19,793  & 65,311    & 8,710    & 25/20/25    \\
OGBN-arxiv       & 169,343 & 1,166,243 & 128      & 15/5/20     \\ \hline
\end{tabular}
}
\label{tab:1}
\end{table}

\subsection{Baselines}

To validate the effectiveness of our proposed strategy CPT, we conduct experiments with the following baseline methods to examine performance:

\begin{itemize}
    \item \textbf{Meta-GNN} \cite{Fan_Chengtai_Kunpeng_Goce_Ting_Ji_2019}: integrates attributed networks with MAML \cite{Finn_Abbeel_Levine_2017} using GNNs.

    \item \textbf{AMM-GNN} \cite{Wang_Luo_Ding_Zhang_Li_Zheng_2020}: AMM-GNN suggests enhancing MAML through an attribute matching mechanism. In particular, node embeddings are adaptively modified based on the embeddings of nodes across the entire meta-task.

    \item \textbf{G-Meta} \cite{Huang_Zitnik_2020}: G-meta employs subgraphs to produce node representations, establishing it as a scalable and inductive meta-learning technique for graphs.

    \item \textbf{TENT} \cite{wang2022task}: TENT investigate the limitations of existing few-shot node classification methods from the lens of task variance and develop a novel task-adaptive framework that include node-level, class-level and task-level.
    
\end{itemize}

\subsection{Implementation Details}

During training, we sample a certain number of meta-training tasks from training classes (i.e., base classes) and train the model with these meta-tasks. Then we evaluate the model based on a series of randomly sampled meta-test tasks from test classes (i.e., novel classes). For consistency, the class splitting is identical for all baseline methods. Then the final result of the average classification accuracy is obtained based on these meta-test tasks. 

Our methodology is formulated utilizing the PyTorch\footnote{https://pytorch.org/} framework, and all subsequent evaluations are executed on a Linux-based server equipped with 1 NVIDIA Tesla V100 GPUs. 
 For the specific implementation setting, we set the number of training epochs $T$ as 2000, the weight decay rate is set as 0.0005, the loss weight $\gamma$ is set as 1, the graph layers are set as 2 and the hidden size is set as in $\left\{ 16,32 \right\}$. The READOUT function is implemented as mean-pooling. The other optimal hyper-parameters were determined via grid search on the validation set: the learning rate was searched in $\left\{ 0.03, 0.01, 0.005 \right\}$, the dropout in $\left\{ 0.2, 0.5, 0.8 \right\}$. Furthermore, to keep consistency, the test tasks are identical for all baselines.

\subsection{Evaluation Methodology}
Following previous few-shot node classification works, we use the average classification accuracy as the evaluation metric and adopt these four few-shot node classification tasks to evaluate the performance of all algorithms: 5-way 3-shot, 5-way 5-shot, 10-way 3-shot, and 10-way 5-shot. The average classification accuracy is the mean result of five complete runs of the model. To further ensure the reliability of the experiment, the model's test performance is repeated ten times, and the average value is taken.

\subsection{Overall Evaluation Results}

\begin{table*}[ht]
    \centering
    \vspace{1mm}
    \caption{Few-shot node classification performance on the four datasets.}
    \vspace{-1mm}
    \resizebox{1.0\textwidth}{!}{
    \begin{tabular}{ccccc|cccc}
    \toprule
    \multirow{2}{*}{Model} & \multicolumn{4}{c}{Cora-full} & \multicolumn{4}{c}{DBLP}  \\
    \cmidrule(l){2-9} 
    & \multicolumn{1}{c}{5-way 3-shot} & \multicolumn{1}{c}{5-way 5-shot} & \multicolumn{1}{c}{10-way 3-shot} & \multicolumn{1}{c}{10-way 5-shot} & \multicolumn{1}{c}{5-way 3-shot} & \multicolumn{1}{c}{5-way-5 shot} & \multicolumn{1}{c}{10-way 3-shot} & \multicolumn{1}{c}{10-way 5-shot} \\ \midrule
    Meta-GNN & 57.60 & 60.92 & 38.48 & 40.68 & 78.38 & 79.34 & 67.02 & 68.54 \\
    Meta-GNN+CPT & 59.40 & 62.36 & 42.45 & 43.81 & 79.68 & 81.60 & 68.84 & 70.24 \\
    $\Delta$ Gain & +3.1\% & +2.4\% & +10.3\% & +7.7\% & +1.7\% & +2.8\% & +2.7\% & +2.5\% \\  
    \midrule
    G-META & 57.93 & 60.30 & 45.67 & 47.76 & 75.49 & 76.38 & 57.96 & 61.18 \\
    G-META+CPT & 59.13 & 62.16 & 46.45 & 49.02 & 76.37 & 76.75 & 60.06 & 63.64 \\
    $\Delta$ Gain & +2.1\% & +3.0\% & +1.7\% & +2.6\% & +1.2\% & +0.5\% & +3.6\% & \textbf{+4.0\%} \\  
    \midrule
    AMM-GNN & 59.06 & 61.66 & 20.82 & 27.98 & 79.12 & 82.34 & 65.93 & 68.36\\
    AMM-GNN+CPT & 63.58 & 64.04 & 30.25 & 40.31 & 81.54 & 84.52 & 68.26 & 68.83 \\
    $\Delta$ Gain & \textbf{+7.7\%} & \textbf{+3.9\%} & \textbf{+45.3\%} & \textbf{+44.1\%} & +3.1\% & +2.6\% & +3.5\% & +0.7\% \\  
    \midrule
    TENT & 64.80 & 69.24 & 51.73 & 56.00 & 75.76 & 79.38 & 67.59 & 69.77 \\ 
    TENT+CPT & 67.48 & 70.50 & 52.99 & 58.82  & 81.40 & 82.98 & 70.51 & 70.89 \\  
    $\Delta$ Gain & +4.1\% & +1.8\% & +2.4\% & +5.1\% & \textbf{+7.4\%} & \textbf{+4.5\%} & \textbf{+4.3\%} & +1.6\% \\  
    \midrule
    Averaged $\Delta$ Gain & +4.3\% & +2.8\% & +14.8\% & +14.9\% & +3.4\% & +2.6\% & +3.5\% & +2.2\% \\  
    \bottomrule
    \end{tabular}
    }
    \resizebox{1.0\textwidth}{!}{
    \begin{tabular}{ccccc|cccc}
    \toprule
    \multirow{2}{*}{Model} & \multicolumn{4}{c}{Amazon-E} & \multicolumn{4}{c}{OGBN-arxiv}  \\
    \cmidrule(l){2-9} 
    & \multicolumn{1}{c}{5-way 3-shot} & \multicolumn{1}{c}{5-way 5-shot} & \multicolumn{1}{c}{10-way 3-shot} & \multicolumn{1}{c}{10-way 5-shot} & \multicolumn{1}{c}{5-way 3-shot} & \multicolumn{1}{c}{5-way-5 shot} & \multicolumn{1}{c}{10-way 3-shot} & \multicolumn{1}{c}{10-way 5-shot} \\ \midrule
    Meta-GNN & 70.26 & 75.10 & 61.14 & 66.01 & 48.22 & 50.24 & 31.30 & 31.98 \\
    Meta-GNN+CPT & 72.32 & 78.32 & 63.95 & 67.58 & 49.18 & 51.92 & 33.87 & 35.11 \\
    $\Delta$ Gain & +2.9\% & +4.3\% & +4.6\% & +2.4\% & +2.0\% & +3.3\% & +8.2\% & +9.8\% \\ 
    \midrule
    G-META & 73.50 & 77.02 & 61.85 & 62.93 & 44.29 & 48.72 & 41.26 & 46.64 \\
    G-META+CPT & 74.64 & 78.26 & 62.72 & 67.07 & 47.28 & 51.08 & 42.24 & 48.68 \\
    $\Delta$ Gain & +1.5\% & +1.6\% & +1.4\% & \bf+6.6\% & \bf+6.8\% & +4.8\% & +2.4\% & +4.3\% \\ 
    \midrule
    AMM-GNN & 73.95 & 76.10 & 62.91 & 68.34 & 50.54 & 52.44 & 30.96 & 32.66 \\
    AMM-GNN+CPT & 76.54 & 78.64 & 67.57 & 69.74 & 52.58 & 56.44 & 32.87 & 35.76 \\ 
    $\Delta$ Gain & +3.5\% & +3.3\% & \bf+7.4\% & +2.0\% & +4.0\% & \bf+7.6\% & +6.2\% & +9.5\% \\ 
    \midrule
    TENT & 74.26 & 78.12 & 65.66 & 70.49 & 55.62 & 62.96 & 41.13 & 46.02 \\ 
    TENT+CPT & 79.32 & 83.98 & 70.25 & 74.35 & 59.24 & 66.56 & 57.64 & 64.50 \\ 
    $\Delta$ Gain & \bf+6.8\% & \bf+7.5\% & +7.0\% & +5.4\% & +6.5\% & +4.1\% & \bf+41.8\% & \bf+40.2\% \\  
    \midrule
    Averaged $\Delta$ Gain & +3.7\% & +4.2\% & +5.1\% & +4.1\% & +4.8\% & +4.9\% & +14.6\% & +15.9\% \\  
    \bottomrule
    \end{tabular}
    }
    \label{tab:result}
    \vspace{-3mm}
\end{table*}

We compare the few-shot node classification performance of four baseline methods and their CPT-enhanced versions in Table \ref{tab:result}. This comprehensive comparison allows us to make the following observations:
\begin{itemize}
    \item CPT brings performance improvement to all baseline methods, demonstrating that CPT technology can effectively enhance the performance of few-shot node classification models. Moreover, the average gain exceeds 3\% in most cases, which is a significant improvement. The gain effect brought by CPT varies for different models, which may be related to the structure and characteristics of the model. Among them, TENT+CPT achieves the highest performance in most cases. 
    \item The 10-way task setting is more challenging than the 5-way setting because it involves more categories. We find that the cases with very high gains all come from 10-way tasks, such as AMM-GNN on Cora-full and TENT on OGBN-arxiv. This indicates that CPT may be more effective for more complex classification tasks.
    \item The 3-shot task setting is more challenging than the 5-shot setting because it has fewer samples to learn from. However, based on the experimental results, the difference in gains between 3-shot and 5-shot tasks is not significant, which suggests that the effect of CPT may not be strongly related to the number of samples.
\end{itemize}

%

\subsection{Ablation Study}

In this section, we conducted an ablation study to verify the effectiveness of the motivation in CPT. Firstly, we remove the second stage from CPT and retained only the first stage, using the default meta-learning strategy to train until convergence. We refer to this variant as CPT w/o -SS. Secondly, we remove the first stage and keep the second stage, directly using the initial model to learn gradually difficult tasks. This variant is referred to as CPT w/o -FS. Finally, we reverse the order of the first and second stages in CPT. In this variant, we first learn hard tasks and then learn easy tasks, and we refer to it as CPT-reverse. The specific experimental results are shown in Table \ref{tab:ablation}. Through the results of the ablation experiments, we have made the following findings:

\begin{itemize}
    \item  We find that the performance of the variant CPT w/o -FS is somewhat better than that of the variant CPT w/o -SS. This indicates that the capability-progressive training strategy can help the model converge more effectively compared to the default episodic meta-learning.
    \item We also find that the performance of the aforementioned two variants is significantly lower than the experimental results of CPT. To understand this, we observe their training process and found that both variants might fall into suboptimal solutions, as depicted in Figure \ref{fig:5-3_recon_loss}. From this, we can observe that although the validation loss curve of the CPT w/o -FS variant is slightly better than that of the CPT w/o -SS variant, overall, their curve trajectories and loss values are very similar. The possible reason for this similarity is that both variants involve random task sampling, which might lead to encountering very challenging tasks early in training. This difficulty in knowledge acquisition could consequently result in convergence towards suboptimal points. In contrast, CPT effectively tackles this problem through its two-stage training approach.
    \item Moreover, we find that if we first learn some difficult tasks and then learn easy tasks, i.e., CPT-reverse, the effect is also worse than CPT. From the perspective of experimental results, the order of learning is important, and the order of learning from easy to hard can help the model learn better. 
\end{itemize}

\begin{table}[h]
    \vspace{-1mm}
    \centering
    \caption{Ablation study with TENT on the Amazon-E.}
    \resizebox{0.47\textwidth}{!}{
    \small
    \begin{tabular}{ccccc}
        \toprule
         Variant & 5way-3shot & 5way-5shot & 10way-3shot & 10way-5shot  \\ \midrule
         CPT w/o -SS & 74.26 & 78.12 & 65.66 & 70.49 \\
         CPT w/o -FS & 74.78 & 80.61 & 67.23 & 71.64 \\   
         CPT-reverse & 76.09 & 81.14 & 67.86 & 72.26 \\ 
         CPT &  79.32 & 83.98 & 70.25 & 74.35  \\ 
        \bottomrule
    \end{tabular}
    \label{tab:ablation}
    }
\end{table}

These phenomena indicate that models have different level competencies at different stages, and the learning process from easy to hard can enhance the model's generalization ability. Furthermore, the two-stage learning strategy can effectively alleviate the suboptimal solution problem.

\subsection{Effectiveness in Mitigating Suboptimal Solutions.}
\begin{figure}[ht!]
  \centering
  \vspace{-3mm}
  
  \begin{minipage}{0.23\textwidth}
    \centering
    \includegraphics[width=\textwidth]{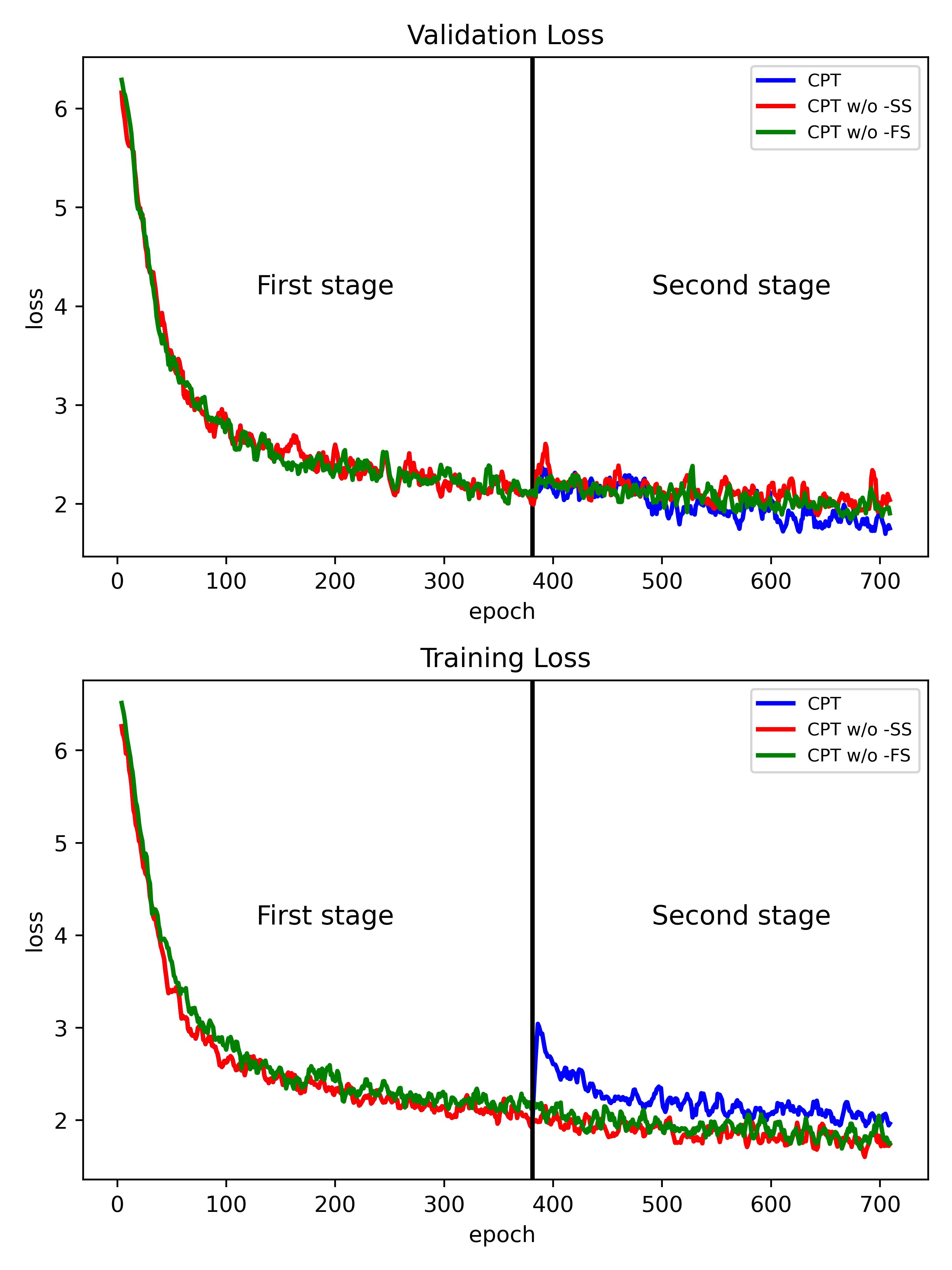}
    \vspace{-5mm}
    \caption{ \textbf{Loss trend comparison of CPT, CPT w/o -SS, and CPT w/o -FS. }}
    \label{fig:5-3_recon_loss}
  \end{minipage}\hfill
  \begin{minipage}{0.23\textwidth}
    \centering
    \includegraphics[width=\textwidth]{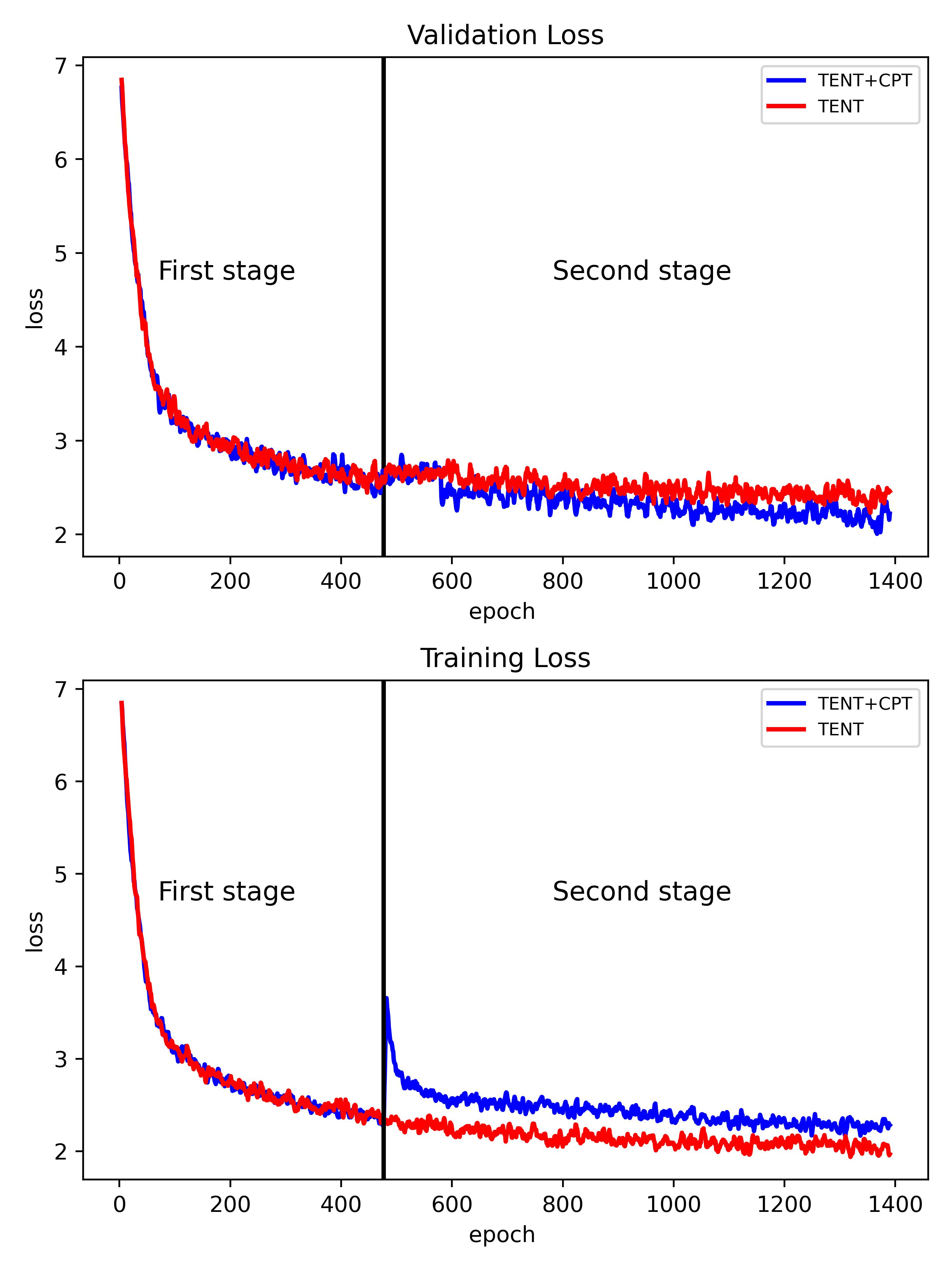}
    \vspace{-5mm}
    \caption{ \textbf{Comparison of the loss trends between TENT and TENT+CPT on Amazon-E.}}
    \label{fig:10-3_recon_loss}

  \end{minipage}
  
  \vspace{-3mm}
\end{figure}

To further verify whether our method can help the model break through suboptimal points, we observe the loss changes of TENT and TENT+CPT, as depicted in Figure \ref{fig:10-3_recon_loss}. In the figure, the loss curve of TENT+CPT is represented in blue, the loss curve of TENT is represented in red, and the black vertical line represents the boundary between the first and second stages. From this observation, we can infer the following:

In the first stage, the training loss and validation loss of the model are essentially consistent, as they follow the same training setup. However, in the second stage, TENT+CPT faces more challenging tasks, causing the training loss to remain relatively high. Despite this, its validation loss gradually decreases and becomes lower than TENT's validation loss. In contrast, TENT exhibits low training loss but high validation loss. This suggests that by applying the CPT technology, the model's generalization ability is improved, further mitigating the suboptimal solution problem.

CPT introduces a certain level of difficulty to the training tasks, pushing the model to learn more general and robust features, which eventually leads to better performance in the validation phase. This outcome supports the idea that the CPT technology  can effectively address the suboptimal solution problem and improve the model's performance in few-shot node classification tasks.

\subsection{Effect of Meta-training Support Set Size}
\begin{figure}[hb]
  \vspace{-3mm}
  \includegraphics[width=0.5\textwidth]{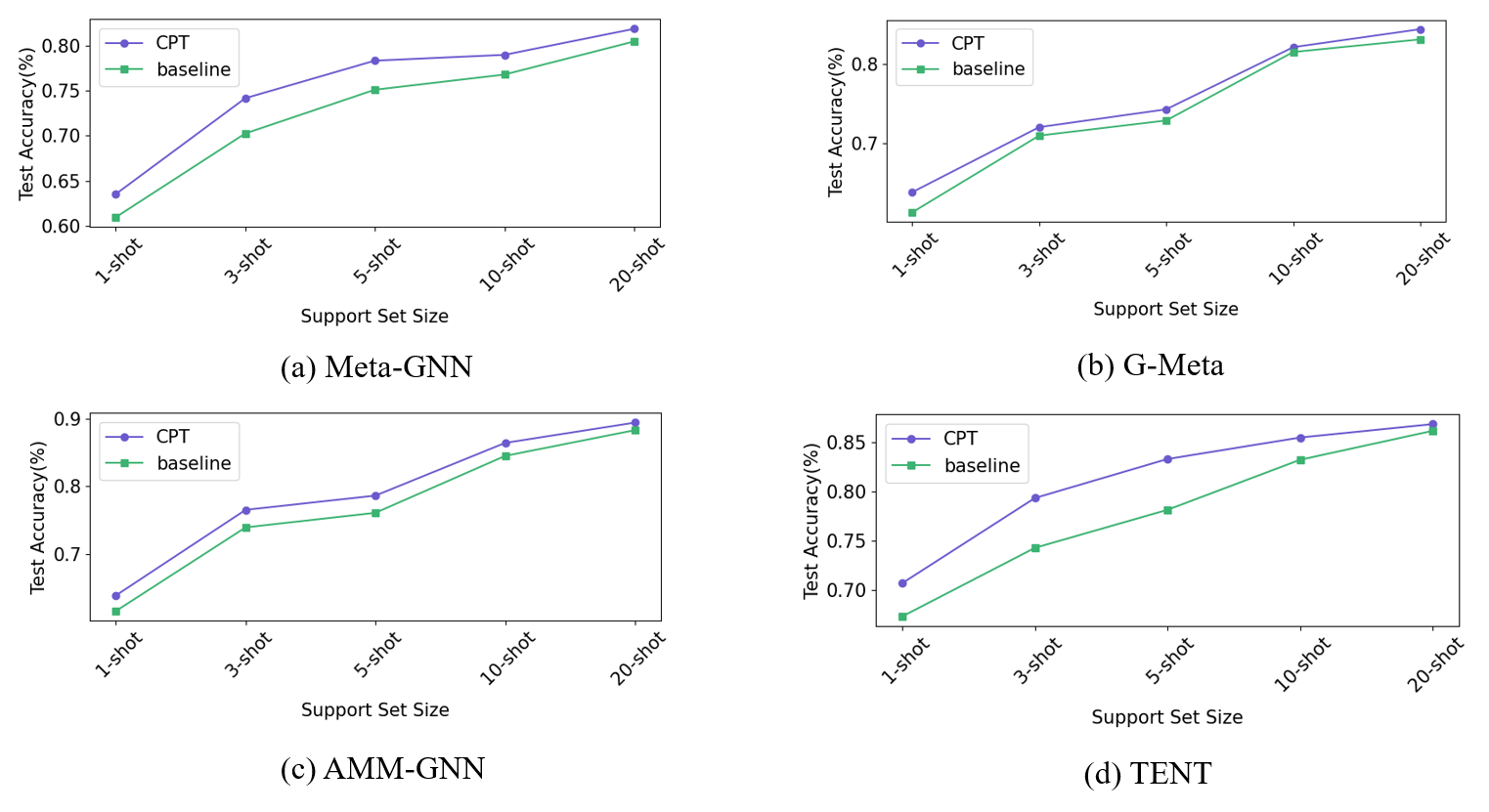}
  \vspace{-5mm}
  \caption{ \textbf{The trend of performance changes for different
baselines and CPT with the increase of support set size
$St$. }}
  \label{fig:S_E}
  \vspace{-3mm}
\end{figure}

we are considering adding some experimental analyses, such as the impact of the size of the support set
on performance. As the support set size $St$ increases, our
method continues to be effective, as shown in Figure \ref{fig:S_E}. We
observe that under different $St$ settings, our method consistently outperforms all baselines.

\section{Related Work}

\subsection{Few-shot Learning on Graphs}
Few-shot Learning (FSL) aims to apply knowledge from tasks with abundant supervision to new tasks with limited labeled data. FSL methods generally fall into two categories: metric-based approaches \cite{Sung_Yang_Zhang_Xiang_Torr_Hospedales_2018,Liu_Zhou_Long_Jiang_Zhang_2019}, like Prototypical Networks \cite{Snell_Swersky_Zemel_2017}, which measure similarity between new instances and support examples, and meta-optimizer-based approaches \cite{Ravi_Larochelle_2017,Mishra_Rohaninejad_Chen_Abbeel_2017}, like MAML \cite{Finn_Abbeel_Levine_2017}, focusing on optimizing model parameters with limited examples.
In the field of graphs, there have been several recent works proposing the application of few-shot learning to graph-based tasks \cite{Ma_Bu_Yang_Zhang_Yao_Yu_Zhou_Yan_2020,Chauhan_Nathani_Kaul_2020}, and they have achieved notable success, particularly in the context of attributed networks.
Among these works, GPN \cite{Ding_Wang_Li_Shu_Liu_Liu_2020} introduces the utilization of node importance derived from Prototypical Networks \cite{Snell_Swersky_Zemel_2017}, leading to improved performance. On the other hand, G-Meta \cite{Huang_Zitnik_2020} employs local subgraphs to learn node representations and enhances model generalization through meta-learning.
TENT \cite{wang2022task} leverages both local subgraphs and prototypes to adapt to tasks, effectively addressing the task-variance issue. It considers three perspectives: node-level, class-level, and task-level.

Although there has been some progress in few-shot learning on graphs, they all share a common issue: the use of random sampling without considering the distribution of task difficulty, which can lead to inefficient knowledge acquisition by the meta-learner. To address this, we propose a task scheduler based on curriculum learning that progresses from easy to difficult tasks, further enhancing the model's knowledge acquisition efficiency.

\subsection{Curriculum Learning}
 
Inspired by the human learning process, Curriculum learning(CL) aims to adopt a meaningful learning sequence, such as from easy to hard patterns\cite{Bengio_Louradour_Collobert_Weston_2009}.
As a general and flexible plug-in, the CL strategy has proven its ability to improve model performance, generalization, robustness, and convergence in various scenarios, including image classification \cite{Cascante-Bonilla_Tan_Qi_Ordonez_2022,Zhou_Wang_Bilmes_2020} , semantic segmentation \cite{Dai_Sakaridis_Hecker_Gool_2019,Feng2020SemiSupervisedSS}, neural machine translation \cite{Guo_Tan_Xu_Qin_Chen_Liu_2020,Platanios_Stretcu_Neubig_Póczos_Mitchell_2019}, etc. The existing CL methods fall into two groups \cite{Li_Wang_Zhu_2023}: (1) Predefined CL \cite{wei2023clnode,hu2022tuneup,Wang_Zhou_Zhao_Wang_Wen_2023} that involves manually designing heuristic-based policies to determine the training order of data samples, and (2) automatic CL \cite{Qu_Tang_Han_2018,Vakil_Amiri_2022,Gu_Zheng_Li_2021} that relies on computable metrics (e.g., the training loss) to dynamically design the curriculum for model training. CL for graph data is an emerging area of research. Recent studies have demonstrated its effectiveness in various graph-related tasks like node classification \cite{wei2023clnode,Qu_Tang_Han_2018}, link prediction \cite{hu2022tuneup,Vakil_Amiri_2022}, and graph classification \cite{Wang_Zhou_Zhao_Wang_Wen_2023,Gu_Zheng_Li_2021}. However, applying CL methods to the challenge of few-shot node classification is still an area with significant hurdles. 

To overcome this, our approach introduces a pioneering training strategy that merges curriculum learning with few-shot learning on graphs, significantly enhancing the model’s performance in few-shot node classification scenarios.

\section{CONCLUSION}

In this paper, we discuss the limitations of the episodic meta-learning approach, where random task sampling might lead the model to converge to suboptimal solutions. To tackle this issue, we propose a novel curriculum-based training strategy called CPT to better train the GNN meta-learner. This strategy is divided into two stages: the first stage trains on simple tasks to acquire preliminary capabilities, and the second stage progressively increases task difficulty as the model's capabilities improve.
We perform thorough experiments using widely-recognized datasets and methods. Our results show that CPT improves all the baseline methods, resulting in a substantial increase in the performance of models for few-shot node classification and offering important insights. In most cases, the average improvement is more than 3\%.
Notably, this is a general  strategy that can be applied to any GNN architecture and loss function.
Therefore, future work can be further extended to other tasks, such as few-shot link prediction and graph classification.

\bibliographystyle{named}
\bibliography{ijcai24}

\end{document}